\documentclass[journal]{IEEEtran}
\usepackage{graphicx}
\usepackage{amsmath}
\usepackage{makecell}
\usepackage{multirow}
\usepackage{collcell}
\usepackage{colortbl}
\usepackage{bm}
\usepackage{tikz}
\usepackage{pgfplots}
\usepackage{pgfplotstable}
\usepackage{enumitem}
\usepackage{adjustbox}
\usepackage{soul}
\usepackage{newfloat}
\usepackage{xcolor}
\usepackage{siunitx}
\usepackage{collcell}
\usepackage{graphicx} 
\usepackage{booktabs}
\usepackage{subcaption}
\usepackage{mathtools}
\usepackage{multirow}
\usepackage{url}
\usepackage{fontawesome}
\usepackage[font=small]{caption}

\usepackage[T1]{fontenc}  
\usepackage{amsmath,amsfonts,amssymb,amsthm}
\DeclareFloatingEnvironment[fileext=dia, within=section]{diagram}
\usepackage[normalem]{ulem}
\useunder{\uline}{\ul}{}
\usepackage{tkz-kiviat}
\usetikzlibrary{arrows}
\usetikzlibrary{patterns}

\usepackage{cleveref}

\usetikzlibrary{arrows,positioning}
\usetikzlibrary{plotmarks}
\usetikzlibrary{calc}
\usepgfplotslibrary{groupplots}
\usetikzlibrary{patterns}
\usetikzlibrary{shapes.geometric}
\pgfdeclarelayer{bg}    
\pgfsetlayers{bg,main, background}
\usepgfplotslibrary{fillbetween}
\pgfplotsset{compat=1.17, compat/show suggested version=false}
\def\Decimal{0.0000}

\def\Ulinehelp#1.#2 {%
  #1.#2\setbox0=\hbox{#1\Decimal}\hspace{-\wd0}{\if\relax#2\relax%
    \uline{\phantom{#1.0}}\else\uline{\phantom{#1.#2}}\fi}%
}
\usetikzlibrary{patterns}

\usetikzlibrary{spy}
\usepackage{algorithm}
\usepackage{algpseudocode}

\ifCLASSINFOpdf
\else
\fi

\definecolor{hotmagenta}{rgb}{1.0, 0.11, 0.81}
\definecolor{gray1}{RGB}{0,0,0}
\definecolor{gray2}{RGB}{80,80,80}
\definecolor{gray3}{RGB}{140,140,140}
\definecolor{gray4}{RGB}{170,170,170}
\definecolor{gray5}{RGB}{200,200,200}
\definecolor{gray6}{RGB}{220,220,220}
\definecolor{coordscolor}{RGB}{235,235,235}

\begin{document}

\title{Segmenting Clouds in Satellite Images Using nnU-Nets}
\title{Hands-Free Cloud Segmentation in Satellite Images Using nnU-Nets}
\title{Cloud Segmentation in Satellite Images Using nnU-Nets}
\title{Self-Configuring nnU-Nets Detect Clouds in Satellite Images}

\author{
Bartosz Grabowski, Maciej Ziaja, Michal Kawulok,~\IEEEmembership{Member,~IEEE}, Nicolas Long\'{e}p\'{e}, Bertrand Le Saux,~\IEEEmembership{Senior Member,~IEEE}, and Jakub Nalepa,~\IEEEmembership{Senior Member,~IEEE}
\thanks{BG, MZ, MK, and JN are with KP Labs, Gliwice, Poland (e-mail: jnalepa@ieee.org). BG is with Institute of Theoretical and Applied Informatics, Polish Academy of Sciences, Gliwice, Poland. MK and JN are with Silesian University of Technology, Gliwice, Poland. NL and BLS are with $\Phi$-lab, European Space Agency, Frascati, Italy. This work was funded by European Space Agency (GENESIS), supported by the ESA $\Phi$-lab ({https://philab.phi.esa.int/}), and by the SUT grant for maintaining and developing research potential. BG acknowledges funding from the Polish budget funds for science in the years 2018–2022, as a scientific project ``Application of transfer learning methods in the problem of hyperspectral images classification using convolutional neural nets'' under the “Diamond Grant” (DI2017 013847).}}

\markboth{}%
{Shell \MakeLowercase{\textit{et al.}}: Bare Demo of IEEEtran.cls for IEEE Journals} 
 
\maketitle

\begin{abstract}

Cloud detection is a pivotal satellite image pre-processing step that can be performed both on the ground and on board a satellite to tag useful images. In the latter case, it can help to reduce the amount of data to downlink by pruning the cloudy areas, or to make a satellite more autonomous through data-driven acquisition re-scheduling of the cloudy areas. We approach this important task with nnU-Nets, a self-reconfigurable framework able to perform meta-learning of a segmentation network over various datasets. Our experiments, performed over Sentinel-2 and Landsat-8 multispectral images revealed that nnU-Nets deliver state-of-the-art cloud segmentation performance without any manual design. Our approach was ranked within the top 7\% best solutions (across 847 participating teams) in the On Cloud N: Cloud Cover Detection Challenge, where we reached the Jaccard index of 0.882 over more than 10k unseen Sentinel-2 image patches (the winners obtained 0.897, whereas the baseline U-Net with the ResNet-34 backbone used as an encoder: 0.817, and the classic Sentinel-2 image thresholding: 0.652).

\end{abstract}


\IEEEpeerreviewmaketitle

\section{Introduction} \label{sec:intro}

Detecting clouds in satellite image data plays a critical role in the pre-processing chain of such imagery~\cite{JEPPESEN2019247}, especially given that the global cloud coverage is about 68\% annually, hence such cloudy areas constitute considerable amount of acquired data~\cite{Li2021}. Understanding which parts of the scene are obscured by clouds can allow us to not only reduce the amount of data for further on-the-ground processing, but may also help us re-schedule the image acquisition process of the particularly important areas to ultimately retrieve clean images. Additionally, the cloud coverage may bring additional information concerning the climate change, hurricanes, volcanic activity and characteristics of volcanic ashes, and many more events that can be observed from space~\cite{38-cloud-1}. Therefore, developing accurate cloud detection algorithms is of paramount importance nowadays to optimize the mission operations through appropriately handling the low-quality or useless (i.e.,~fully covered by clouds) image data~\cite{Mahajan2020}.

Capturing large, representative, and heterogeneous annotated cloud detection sets is cumbersome and difficult in practice, as they should reflect various factors that affect the satellite image characteristics, such as atmospheric distortions, latitude, ground reflectance, and other~\cite{rs13081532}. Despite those challenges, new sets have been emerging---they are commonly created in a semi-automated way with certain quality procedures adopted to ensure sufficient ground-truth (GT) quality. Such benchmarks can be used for training and verifying cloud detection, and they include the cloud masks for Sentinel-2 (S-2) images~\cite{rs13204100}, the S-2 set released in the On Cloud N Challenge\footnote{This set was created as part of a crowdsourcing competition, and later validated using a team of expert annotators. It is available at \url{https://mlhub.earth/data/ref_cloud_cover_detection_challenge_v1} (accessed on June 3, 2022).}, or the Landsat-8 (L-8) 38-Cloud set~\cite{38-cloud-1,38-cloud-2}.

In the classic image analysis cloud detection techniques, we distinguish rule-based algorithms and time differentiation methods~\cite{Mahajan2020}. Although they are time-efficient and trivial to implement, they generalize poorly across different sensors and acquisition conditions, suffer from thin cloud omission and non-cloud bright pixel commission, and are heavily based on the prior knowledge about the cloud characteristics. On the other hand, supervised machine learning techniques benefit from the GT examples containing manually or semi-automatically delineated clouds to train detectors~\cite{10.1117/1.JRS.15.028507}. Classic algorithms require designing feature extractors and selectors, whereas the recent breakthroughs of deep learning allow us to benefit from automated representation learning~\cite{8898628}. Mohajerani and Saeedi designed a fully-convolutional Cloud-Net architecture for L-8 images~\cite{38-cloud-1}. This manually-crafted and specialized architecture significantly outperformed a more generic U-Net network~\cite{38-cloud-2} and the improved Fmask algorithm fine-tuned for L-8~\cite{ZHU2015269} (originally developed for Landsats 4--7~\cite{ZHU201283}). Albeit delivering high-quality cloud masks, those algorithms required manual redesign process to be applicable to L-8 imagery, and would likely need a similar procedure if they were to be deployed for other satellite images, which reduces their flexibility. In a similar vein, Domnich et al.~utilized the U-Nets for detecting clouds in S-2 images~\cite{rs13204100}, and Yanan et al.~enhanced U-Nets with attention modules to decrease the false cloud detections in snowy areas of L-8 imagery~\cite{Yanan_2020}. The multi-scale feature fusion was also exploited in another encoder-decoder architecture proposed in~\cite{LI2019197}, and in the spectral encoder suggested by Francis et al.~\cite{9615070}. To the best of our knowledge, these are the only two works which reported promising results obtained over different data sources. All of the above-mentioned deep learning algorithms were manually designed by humans. This involves selecting their architecture, pre- and post-processing routines, training strategy, and parameterization (such as the number or size of the kernels)---it is commonly a trial-and-error and time-consuming procedure. To effectively tackle it, various automated ML (AutoML) techniques which optimize the network architecture through the network architecture search~\cite{DBLP:conf/gecco/LorenzoN18}, hyperparameters~\cite{DBLP:conf/gecco/LorenzoNKRP17}, or both~\cite{DBLP:journals/corr/abs-2112-09245} were introduced for different tasks, also including satellite image classification~\cite{10.1007/978-3-030-86517-7_28}. However, such AutoML approaches have not been utilized for the fundamental task of cloud detection. We address this research gap to make the deployment of the cloud detection algorithms as flexible and smooth as possible for emerging satellite missions.

We introduce a fully data-driven AutoML-powered approach for cloud detection in satellite image data which requires zero user intervention. We build upon the nnU-Net (which unfolds to no-new-U-Net) framework that has already established the state of the art in an array of medical image analysis tasks, including the segmentation of brain tumors, liver, prostate, spleen or kidney~\cite{isensee_nnu-net_2021}. In the hands-free processing chain, the nnU-Net adapts its most important components according to the input image data, hence the resulting deep learning model, together with the pre- and post-processing routines, data augmentation and training settings are directly influenced by the characteristics of the training set. This opens new doors for deploying such cloud detection engines in emerging applications based solely on the available image data, without the need of manually redesigning the algorithms. Our experiments performed over two datasets captured by different missions (Landsat-8 and Sentinel-2) showed that the nnU-Nets deliver the segmentation quality competitive with the state of the art (we were ranked in the top 7\% teams within the On Cloud N: Cloud Cover Detection Challenge with almost 850 participating teams). Finally, the elaborated architectures, together with their parameterization, and our code for preparing the L-8 and S-2 images for nnU-Nets are available at  \url{https://gitlab.com/jnalepa/nnUNets_for_clouds}.

\section{Materials and Methods}\label{sec:method}

We discuss the data used in our study in Sect.\,\ref{sec:datasets}. To show flexibility, we exploit both S-2 and L-8 images. The nnU-Nets for segmenting clouds are presented in Sect.\,\ref{sec:method}.

\subsection{Description of Datasets}\label{sec:datasets}

The \textbf{38-Cloud (38-C)} dataset contains 38 L-8 images and their pixel-level GT cloud segmentations. The images are cropped into $384\times384$ patches by the dataset's authors~\cite{38-cloud-1,38-cloud-2}. There are 8,400 training and 9,201 test patches. In 38-Cloud, thin clouds (haze) are also annotated as clouds (i.e.,~as well as thick clouds). The \textbf{On Cloud N: Cloud Cover Detection Challenge (OCN)} dataset consists of S-2 satellite imagery divided into 11,748 training image patches, collected between 2018 and 2020, and 10,980 test image patches (all of them are of $512\times 512$ pixels and have 10\,m spatial resolution). Each image patch was captured for a specific area, mostly in Africa, South America, and Australia. Although the organizers of the challenge included four bands in the competition data (B02, B03, B04, and B08), the publicly available S-2 database could be used during the challenge to pull all other missing bands. We, however, exploited the bands suggested by the organizers only, in order to verify the effectiveness of the nnU-Nets in this very experimental scenario. The cloud labels for OCN were generated using human annotation of the optical S-2 bands (the masks were reviewed by expert annotators). The spectral and spatial characteristics of both sets are given in the supplement.

\subsection{Proposed Method}\label{sec:method}

The nnU-Net algorithm is a deep learning segmentation method that automatically adapts itself based on the underlying characteristics of the training data and target segmentation problem. This configuration encompasses data pre-processing, network architecture, training parameters, and basic post-processing. To understand all aspects of the nnU-Net framework, we refer to Fig.~2 in~\cite{isensee_nnu-net_2021}---the authors discussed which model's properties are set based on the dataset, how they are used to infer specific parameters of the pipeline, and how the ensemble of separate models is finally built.


\begin{figure}[ht!]
     \centering
     \includegraphics[width=1\columnwidth]{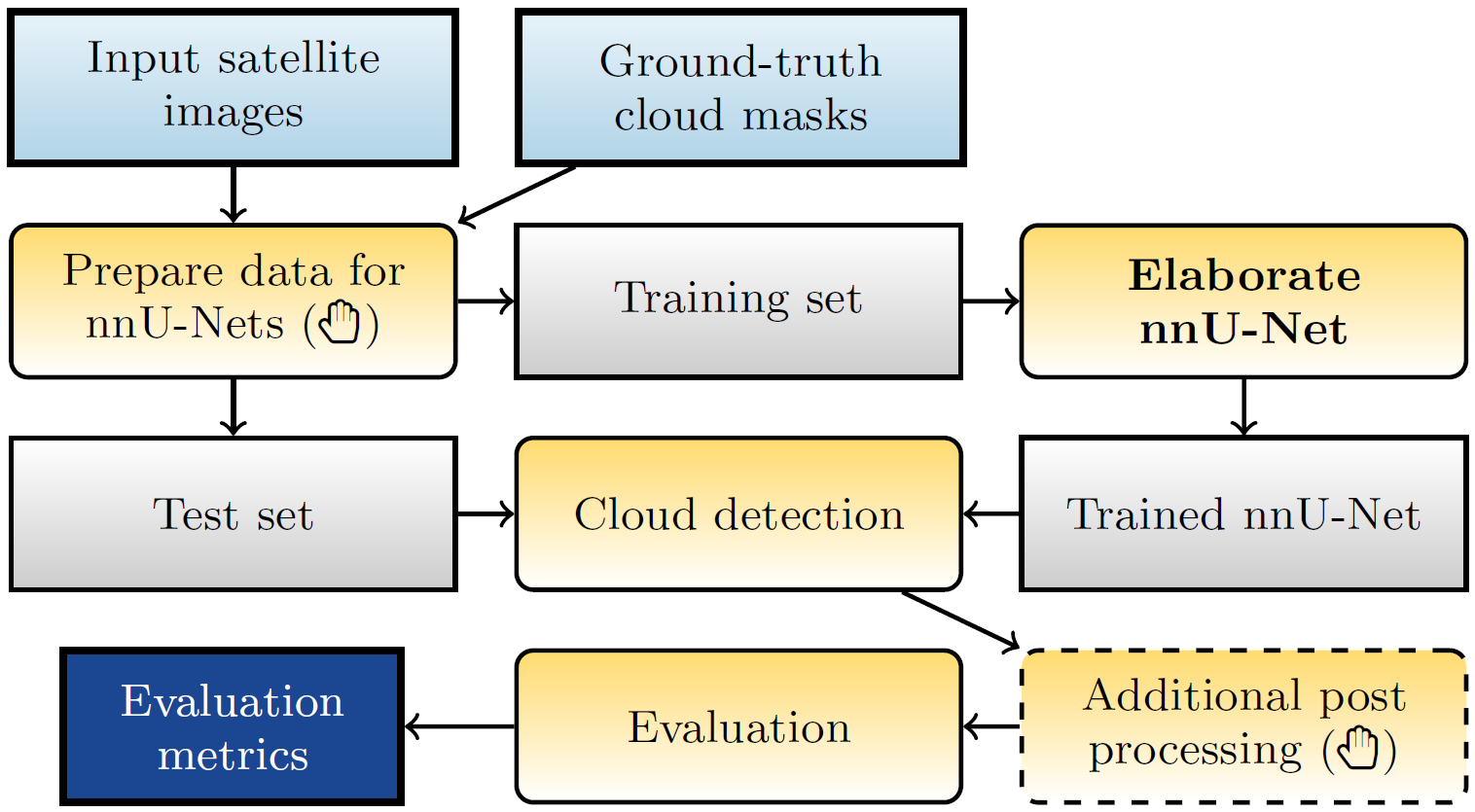}
     \caption{A high-level flowchart presenting an end-to-end deployment pipeline which exploits nnU-Nets. The optional step is rendered as a dashed block, and the manual steps are annotated with \faHandPaperO. In bold, we present the step which is fully automated in nnU-Net, and would have to be performed manually in other non-AutoML approaches.}
     \label{fig:nnunet_flowchart}
 \end{figure}

In Fig.~\ref{fig:nnunet_flowchart}, we present a high-level flowchart rendering an end-to-end deployment pipeline for cloud detection using nnU-Nets. There is only one manual step which needs to be performed manually beforehand, and it involves preparing the image data. Since the nnU-Nets were originally developed for medical image data, they operate on the Neuroimaging Informatics Technology Initiative (NIfTI) files which is an open file format commonly used to store brain imaging data obtained using Magnetic Resonance Imaging methods, and they are widely used in medical image analysis~\cite{DIAZ202125}.

Once the annotated training data is fed to the framework, it is transformed into a standardized dataset representation (it captures e.g.,~the image size or class imbalance ratio), and the nnU-Net re-configures itself to generate an end-to-end segmentation pipeline. The optimization is based on distilling the underlying domain knowledge into the rule-based and empirical hyper-parameters. Afterwards, a set of heuristics is used to model the dependencies between a selected parameter and anticipated network's performance, e.g., ``larger patches increase the contextual information available, thus should be preferred''. The data-related knowledge is distilled into such rules by e.g.,\,initializing the initial patch size to the median image size within the training data, and iteratively decreasing it until the deep network can be trained with some GPU constraints\footnote{The detailed description of the nnU-Net adaptation rules and heuristics is included in the supplementary material of the work by Isensee et al.~\cite{isensee_nnu-net_2021}.}. The hyper-parameters that undergo adaptation encompass, among others, the network topology, patch and batch sizes, normalization, or building an ensemble of base nnU-Net models (including 2D, 3D, and cascaded 3D U-Net-based architectures). Although Isensee et al.~\cite{isensee_nnu-net_2021} claimed that the nnU-Net framework can be utilized in the biomedical domain and indeed showed its superior performance in more than 20 segmentation problems, this approach---to the best of our knowledge---has never been exploited to tackle the Earth observation image analysis tasks. We address this research gap and hypothesize that nnU-Nets can be deployed in a hands-free manner in the satellite image analysis, and that they can deliver competitive performance without any manual intervention. In Fig.~\ref{fig:nnunet_flowchart}, we included an optional post-processing step (rendered as a dashed block)---we will experimentally show (over the OCN dataset) that it can deliver only minimal improvements over fully-automated nnU-Nets (Section~\ref{sec:experiments}).

\section{Experimental Results}\label{sec:experiments}

\newcommand{\Jaccard}{JI}
\newcommand{\Precision}{Pr}
\newcommand{\Recall}{Re}
\newcommand{\Specificity}{Spe}
\newcommand{\OverallAccuracy}{OA}

The objectives of the experimental study are two-fold: to (\textit{i})~confront the nnU-Nets for cloud detection with other state-of-the-art techniques, and to (\textit{ii})~verify their flexibility over satellite imagery captured using different missions. To quantify the performance of our models over the 38-C dataset, we use the Jaccard index (\Jaccard), precision (\Precision), recall (\Recall), specificity (\Specificity) and overall accuracy (\OverallAccuracy), whereas for the OCN we report JI only, as only this metric was calculated by the independent validation server. All results are reported for the test sets (unless stated otherwise) that were unseen during the training process which lasted for 1000 epochs. The experiments ran on an NVIDIA RTX 3090 GPU with 24 GB VRAM, and a single training process took approx. 48 h. The resulting nnU-Net models, together with their entire parameterization are available at \url{https://gitlab.com/jnalepa/nnUNets_for_clouds}, whereas the original nnU-Net framework is accessible at \url{https://github.com/MIC-DKFZ/nnUNet}.

\begin{table}[ht!]
\centering
\scriptsize
  \caption{Comparison of the nnU-Nets with other techniques specifically designed to detect clouds in L-8 imagery. The best metrics are boldfaced, whereas the second best are underlined.}
  \label{tab:results}
  \renewcommand{\tabcolsep}{3.9mm}
  \begin{tabular}{rcccccc}
    \Xhline{2\arrayrulewidth}
    Algorithm & \Jaccard & \Precision & \Recall & \Specificity & \OverallAccuracy\\
    \hline
    FCN$\dagger$~\cite{38-cloud-2} & 0.722 & \uline{0.846} & 0.814 & \uline{0.985} & 0.952\\
    Fmask~\cite{ZHU2015269} & 0.752 & 0.777 & \textbf{0.972} & 0.940 & 0.949\\
    Cloud-Net~\cite{38-cloud-1} & \textbf{0.785} & \textbf{0.912} & 0.849 & \textbf{0.987} & \textbf{0.965}\\
    nnU-Net & \uline{0.756} & {0.845} & \uline{0.866} & 0.976 & \uline{0.953}\\
    \Xhline{2\arrayrulewidth}
    \multicolumn{6}{l}{$\dagger$ Trained with the 38-C training dataset.}
\end{tabular}
\end{table}

The results obtained for the test 38-C patches gathered in Table~\ref{tab:results} indicate that the ensemble of five 3D nnU-Nets (with approx. 25M trainable parameters; processing all 10k test patches took less than one hour) offers competitive performance which is on par with the performance delivered by the methods specifically designed to detect clouds in the L-8 multispectral images. We compared the proposed method with the improved Fmask algorithm that was fine-tuned for L-8 images~\cite{ZHU2015269}, Cloud-Net~\cite{38-cloud-1}, and a more generic fully-convolutional U-Net network trained over the training 38-C data (FCN)~\cite{38-cloud-2}. Although we cannot claim that the nnU-Nets outperformed the hand-crafted techniques, we emphasize that the nnU-Nets were obtained without any manual design or user intervention. The distribution of the metrics shows that the nnU-Nets extracted high-quality cloud masks for the vast majority of scenes (see supplement). In Fig.~\ref{fig:examples}, we present the visual examples (best, worst, and median-quality segmentation as measured by \Jaccard; white areas show clouds), indicating that for mountainous areas the nnU-Nets can incorrectly annotate pixels as clouds (last row). Here, we rendered the full scenes, but the segmentation is performed at the patch level, as the authors of 38-C split the scenes into patches to extract the test set. Due to limited contextual information, it was impossible to precisely segment the most challenging parts of the scenes (middle row). The patch-wise processing leads to the artifacts at the patch boundaries (first row)---it can be mitigated by using the full scenes (or larger patches, depending on the GPU memory) for training and inference.

\begin{figure}[ht!]
\centering
\scriptsize
\renewcommand{\tabcolsep}{0.2mm}
\newcommand{\mymagnification}{6}
\newcommand{\mywidth}{0.29}
\begin{tabular}{cccc}
\Xhline{2\arrayrulewidth}
     & RGB & Ground truth & nnU-Net\\
     \hline
     \rotatebox{90}{~Best (\Jaccard: 0.985)} & \begin{tikzpicture}
        [,spy using outlines={circle,pink,magnification=\mymagnification,size=1.5cm, connect spies}]
        \node {\pgfimage[width=\mywidth\columnwidth]{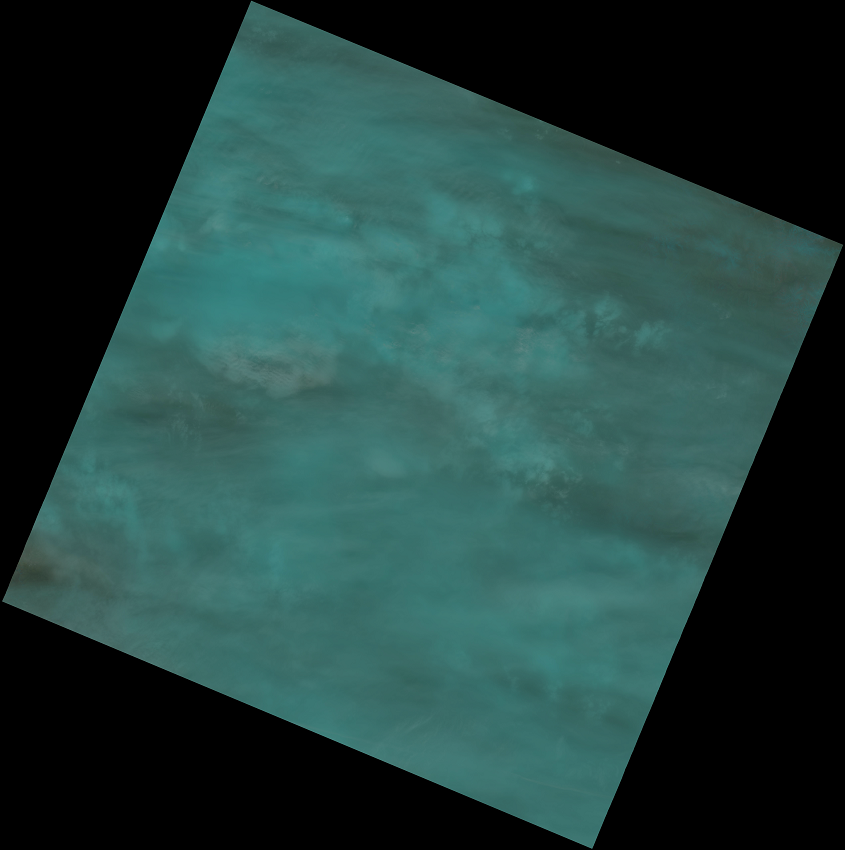}};
        \spy[every spy on node/.append style={thick},every spy in node/.append style={thick}] on (1.1,0.527) in node [left] at (1.1,-0.5);
        \end{tikzpicture}& \begin{tikzpicture}
        [,spy using outlines={circle,pink,magnification=\mymagnification,size=1.5cm, connect spies}]
        \node {\pgfimage[width=\mywidth\columnwidth]{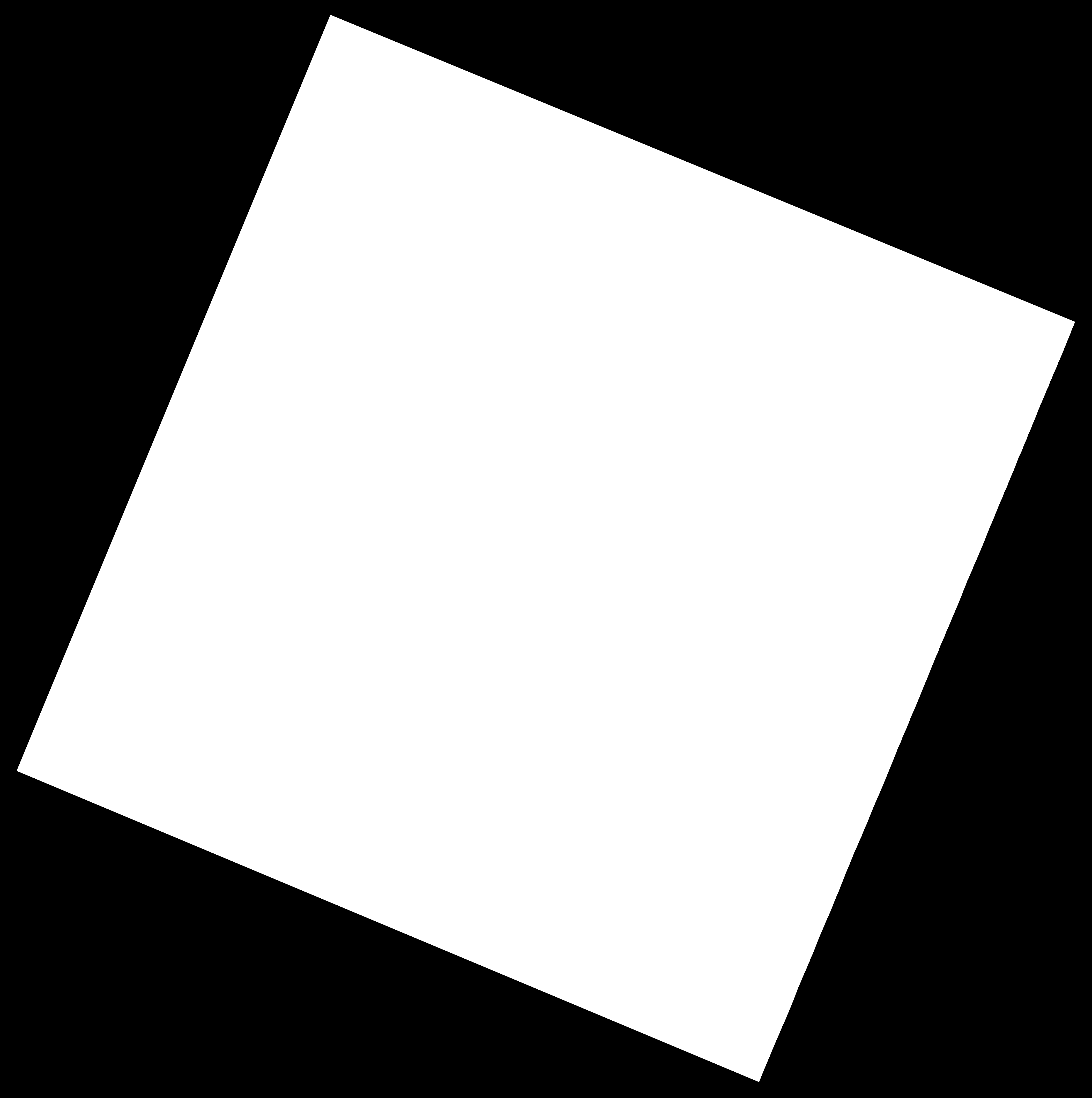}};
        \spy[every spy on node/.append style={thick},every spy in node/.append style={thick}] on (1.1,0.5) in node [left] at (1.1,-0.5);
        \end{tikzpicture}& \begin{tikzpicture}
        [,spy using outlines={circle,pink,magnification=\mymagnification,size=1.5cm, connect spies}]
        \node {\pgfimage[width=\mywidth\columnwidth]{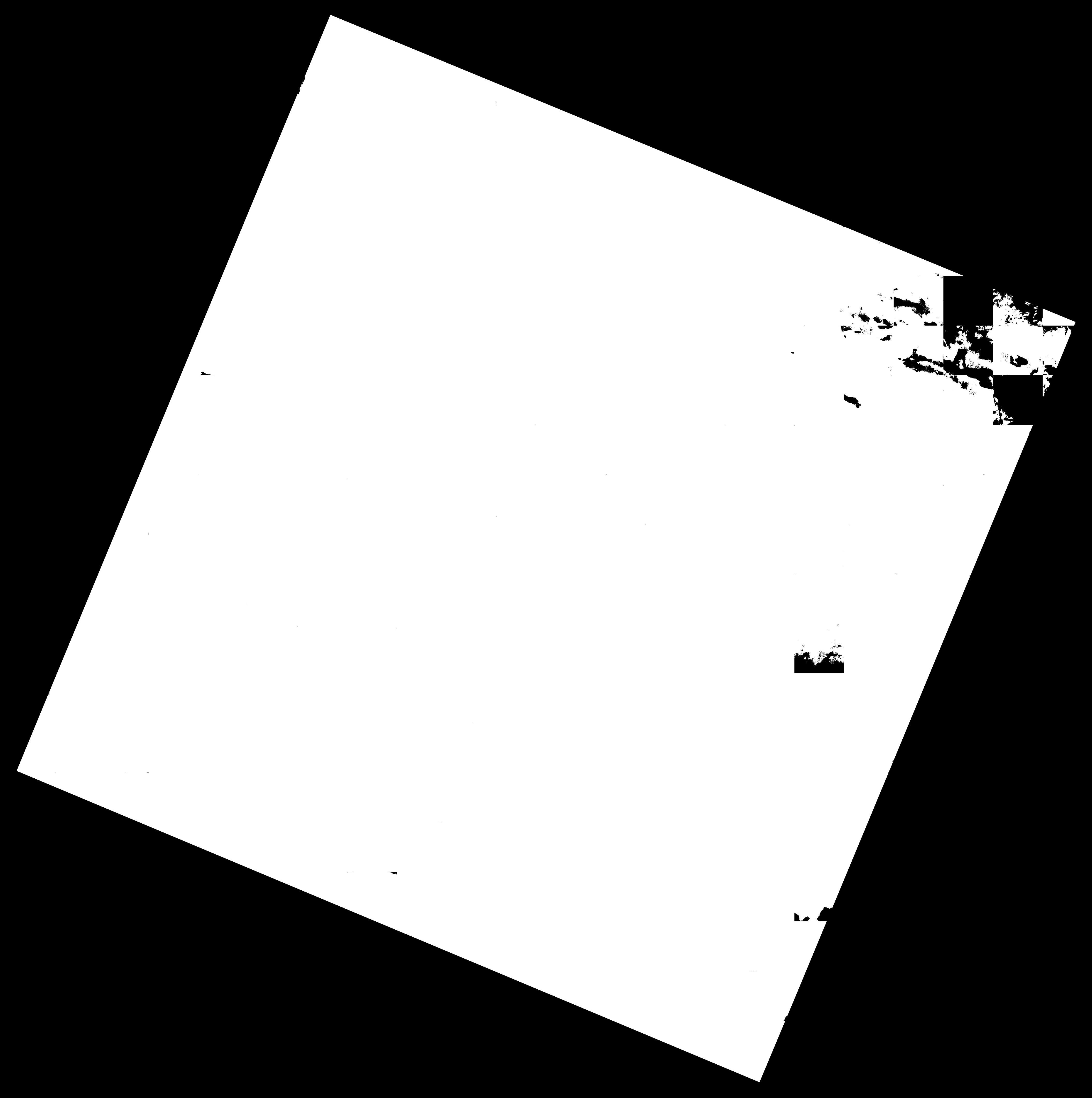}};
        \spy[every spy on node/.append style={thick},every spy in node/.append style={thick}] on (1.1,0.5) in node [left] at (1.1,-0.5);
        \end{tikzpicture}\\[-0.1cm]
     \hline
     \rotatebox{90}{~Median (\Jaccard: 0.806)} &\begin{tikzpicture}
        [,spy using outlines={circle,pink,magnification=\mymagnification,size=1.5cm, connect spies}]
        \node {\pgfimage[width=\mywidth\columnwidth]{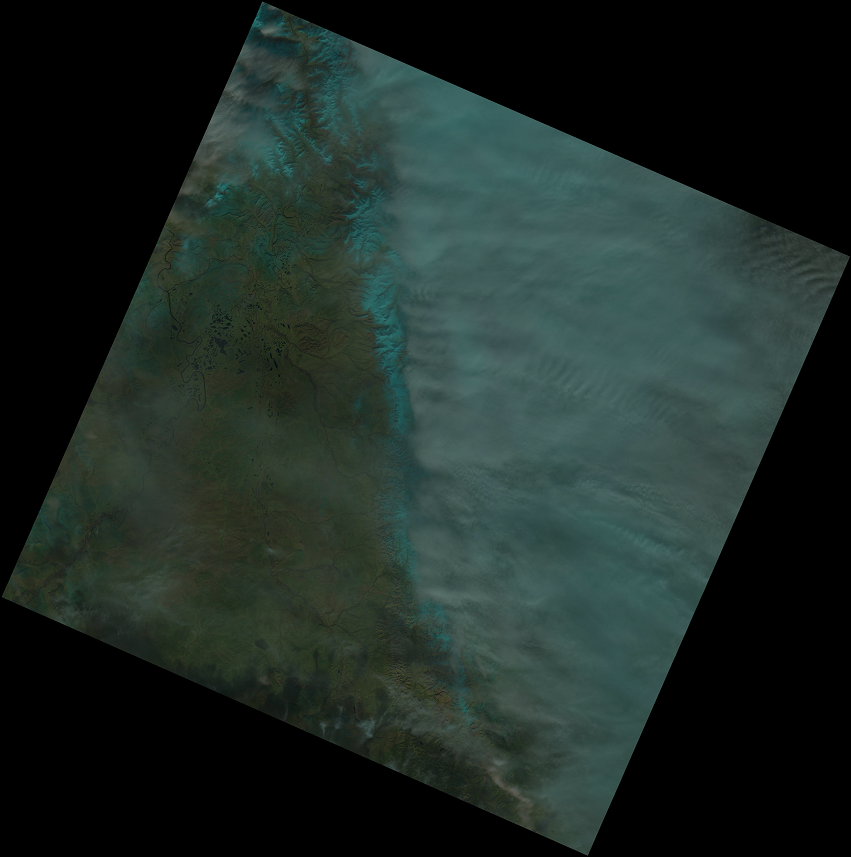}};
        \spy[every spy on node/.append style={thick},every spy in node/.append style={thick}] on (-0.4,0.5) in node [left] at (1.1,-0.5);
        \end{tikzpicture}& \begin{tikzpicture}
        [,spy using outlines={circle,pink,magnification=\mymagnification,size=1.5cm, connect spies}]
        \node {\pgfimage[width=\mywidth\columnwidth]{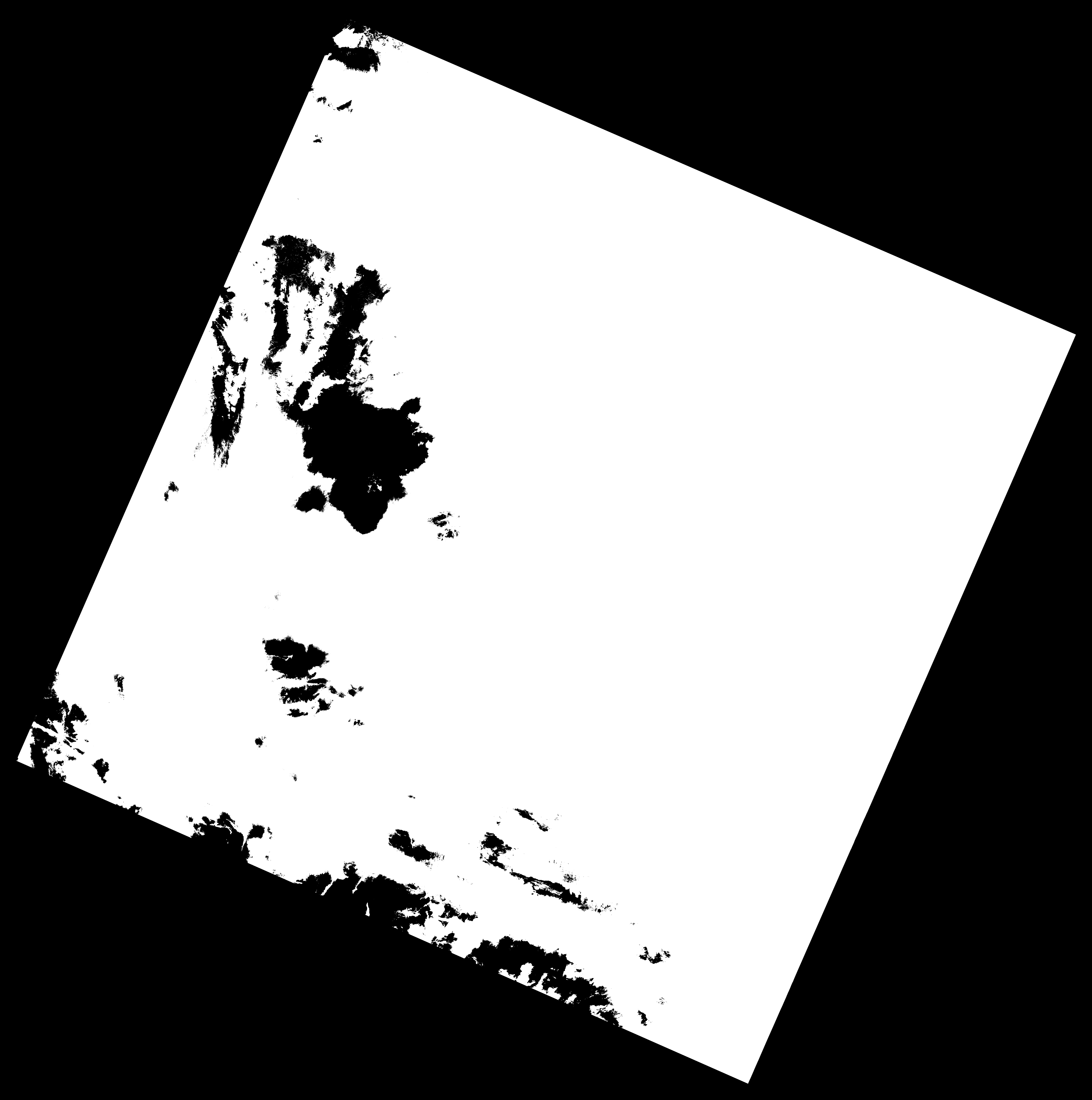}};
        \spy[every spy on node/.append style={thick},every spy in node/.append style={thick}] on (-0.4,0.5) in node [left] at (1.1,-0.5);
        \end{tikzpicture}& \begin{tikzpicture}
        [,spy using outlines={circle,pink,magnification=\mymagnification,size=1.5cm, connect spies}]
        \node {\pgfimage[width=\mywidth\columnwidth]{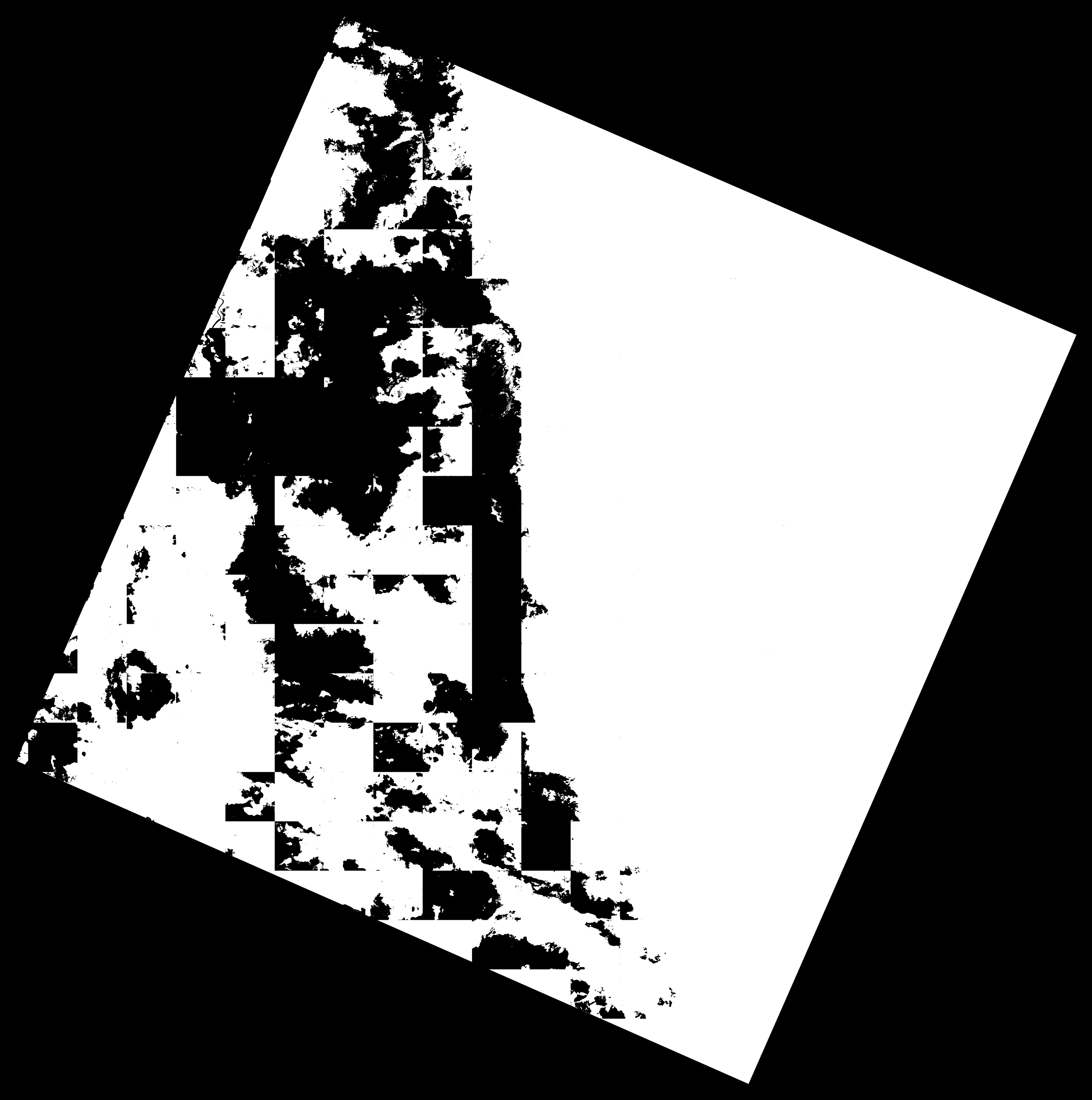}};
        \spy[every spy on node/.append style={thick},every spy in node/.append style={thick}] on (-0.4,0.5) in node [left] at (1.1,-0.5);
        \end{tikzpicture}\\[-0.1cm]
     \hline
     \rotatebox{90}{~Worst (\Jaccard: 0.135)} &\begin{tikzpicture}
        [,spy using outlines={circle,pink,magnification=\mymagnification,size=1.5cm, connect spies}]
        \node {\pgfimage[width=\mywidth\columnwidth]{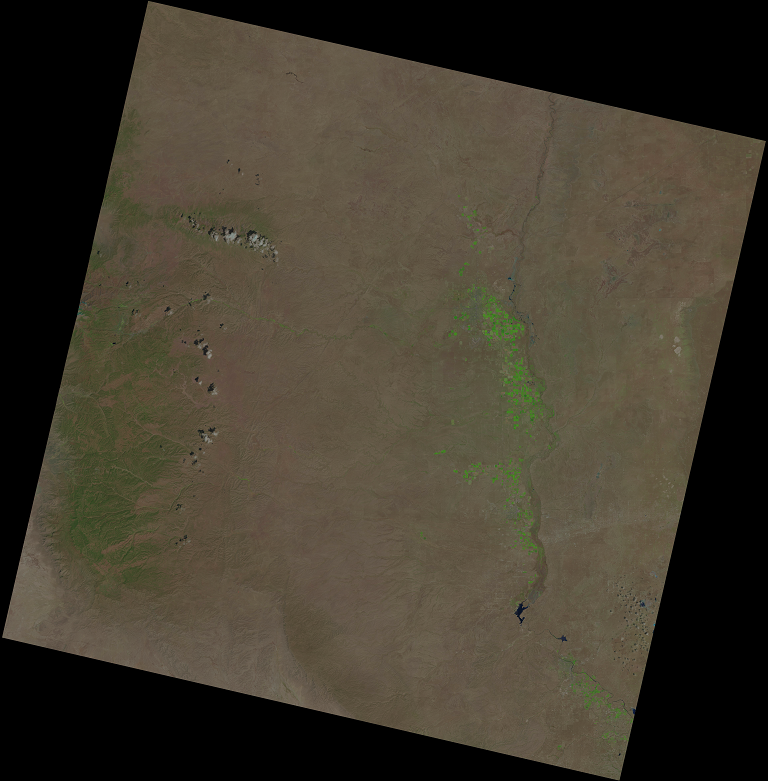}};
        \spy[every spy on node/.append style={thick},every spy in node/.append style={thick}] on (-1.15,-0.4) in node [left] at (1.1,-0.5);
        \end{tikzpicture}&
     \begin{tikzpicture}
        [,spy using outlines={circle,pink,magnification=\mymagnification,size=1.5cm, connect spies}]
        \node {\pgfimage[width=\mywidth\columnwidth]{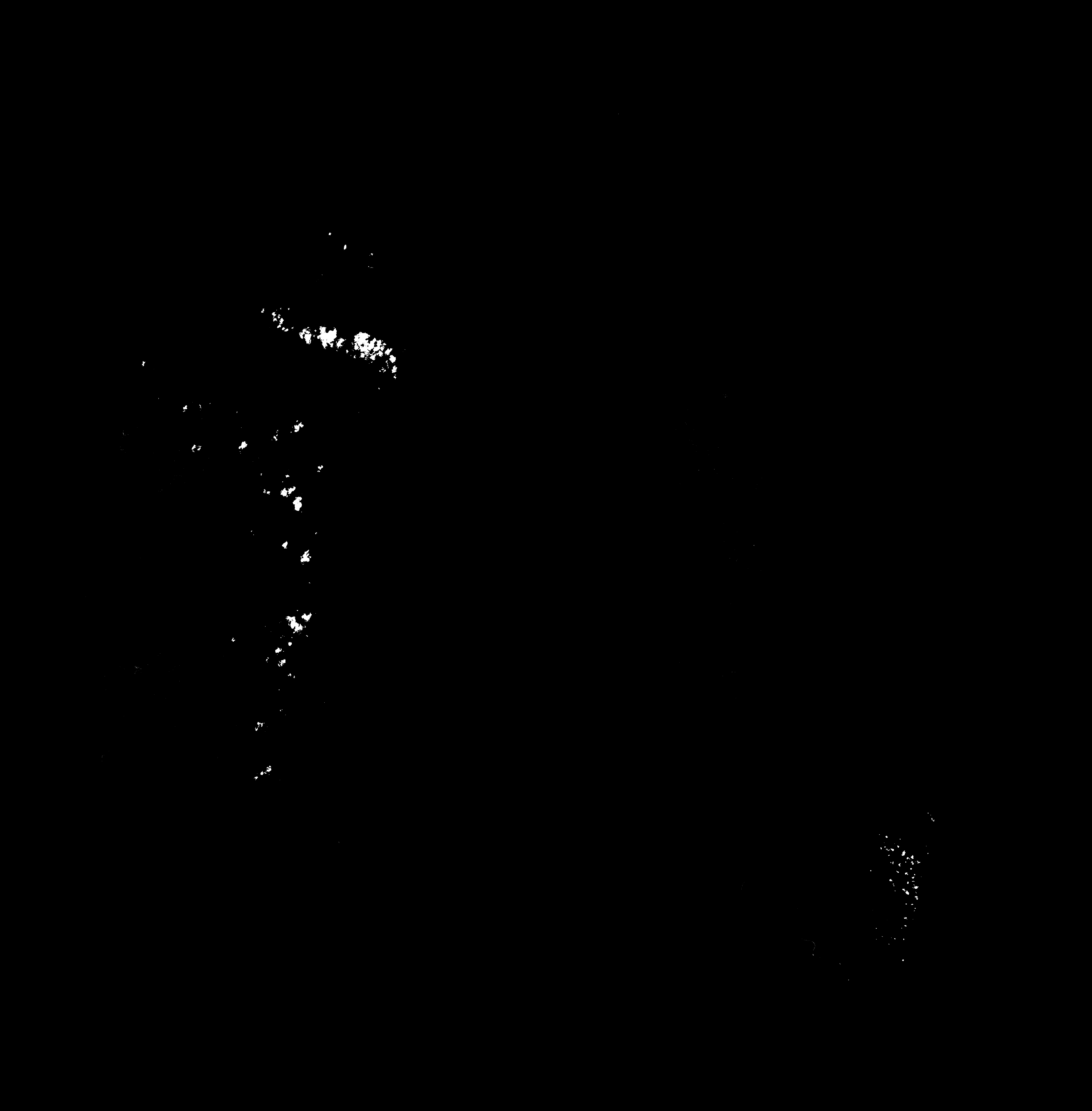}};
        \spy[every spy on node/.append style={thick},every spy in node/.append style={thick}] on (-1.15,-0.4) in node [left] at (1.1,-0.5);
        \end{tikzpicture}&
     \begin{tikzpicture}
        [,spy using outlines={circle,pink,magnification=\mymagnification,size=1.5cm, connect spies}]
        \node {\pgfimage[width=\mywidth\columnwidth]{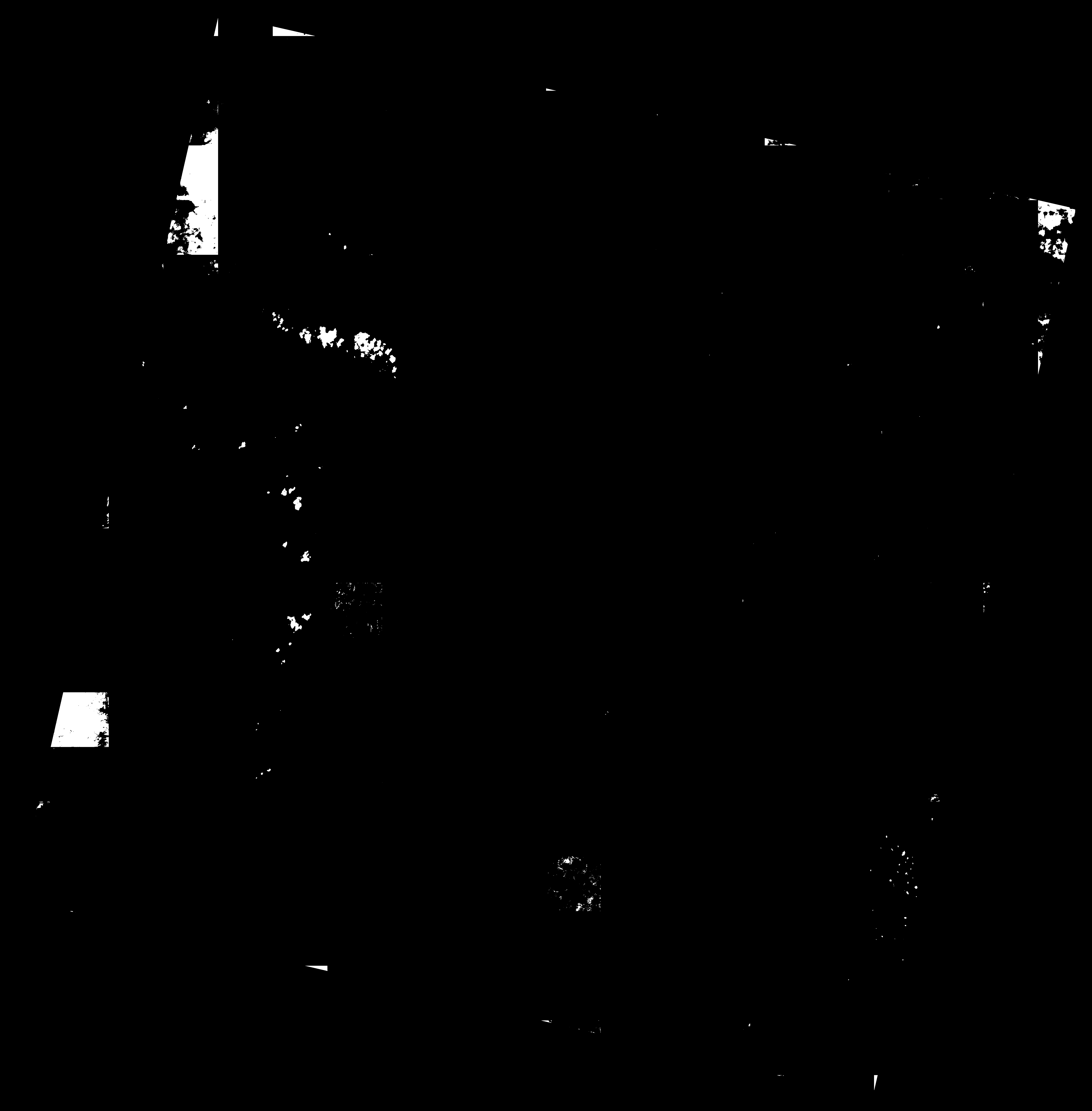}};
        \spy[every spy on node/.append style={thick},every spy in node/.append style={thick}] on (-1.15,-0.4) in node [left] at (1.1,-0.5);
        \end{tikzpicture}\\[-0.1cm]
     \Xhline{2\arrayrulewidth}
\end{tabular}
\caption{Example 38-C cloud masks obtained using the nnU-Nets.}\label{fig:examples}
\end{figure}

To investigate the flexibility of the nnU-Net pipeline, we exploited it for the S-2 data within the On Cloud N: Cloud Cover Detection Challenge. Here, due to the execution time constraints imposed by the organizers (4\,h for processing the entire test set of more than 10k S-2 patches), we exploited an ensemble of four (instead of five) 3D nnU-Nets (approx. 30M trainable parameters). In this experiment, we additionally employed a manually-designed heuristic post processing routine based on the morphological alterations of the resulting nnU-Net cloud masks. In the nnU-Net with our post processing, we execute the closing operation (with the $3\times 3$ kernel) if more than 50\% of pixels within the patch are annotated as clouds in the segmentation map (otherwise, we apply the opening operation with the same kernel). This approach was hand-crafted based on the visual inspection of the results obtained for both training and test S-2 patches (for the latter, we did not have the GT). We observed that there were patches with small false-positive objects (especially in the snowy areas), whereas for the cloudy scenes, we noticed under-segmentation.

\begin{figure}[ht!]
\centering
\includegraphics[width=0.9\columnwidth]{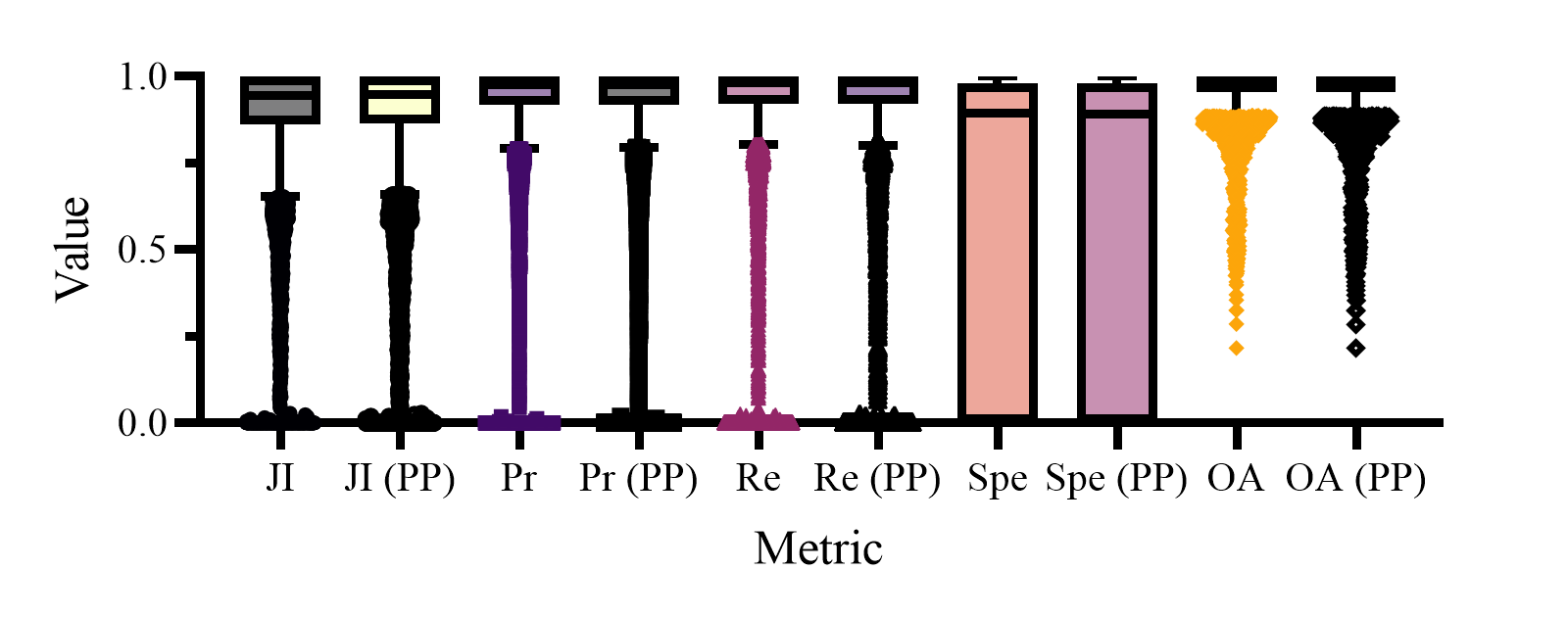}\\[-0.3cm]
\caption{The Tukey box plots showing the distribution of the metrics obtained for approx. 12k \textit{training} S-2 patches from the OCN set using the nnU-Nets with (PP) and without manually-designed post processing. The individual points present outliers (values lower than the 25$^{\rm th}$ percentile of the metric minus 1.5 inter-quartile distance). }\label{fig:s2_tukey}
\end{figure}

Although the differences (Fig.~\ref{fig:s2_tukey}) between the nnU-Net models with and without post processing are statistically significant in the pairwise comparisons over the training data (Wilcoxon two-tailed tests, $p<0.05$), those improvements only marginally affect the scores obtained over the test data (\Jaccard\, increased from 0.8822 to 0.8824). It indicates that the nnU-Net framework automatically elaborates well-performing models which are extremely challenging to improve manually. Overall, we have investigated 18 variants of the nnU-Nets with various post processing, updated loss functions, and without and with ensembling (in the former approach, we trained a single 2D/3D model over the entire training set, whereas in the latter, four deep models are trained independently over non-overlapping folds, and then are fused within the nnU-Net), and none of them delivered further improvements in \Jaccard.

\begin{table}[ht!]
\centering
\scriptsize
  \caption{Comparison of the nnU-Nets with other techniques introduced within the On Cloud N: Cloud Cover Detection Challenge. The result (average \Jaccard) obtained using the nnU-Nets is boldfaced.}
  \label{tab:results_challenge}
  \renewcommand{\tabcolsep}{1mm}
  \begin{tabular}{rcccccccccc}
    \Xhline{2\arrayrulewidth}
    Rank & 1 & 2 & 3 & 10 & 20 & 30 & 40 & \textbf{53} & 107 & S-2\\
    \hline
    \Jaccard & 0.897 & 0.897 & 0.896 & 0.895 & 0.894 & 0.892 & 0.889 & \textbf{0.882} & 0.815 & 0.652\\
    \Xhline{2\arrayrulewidth}
    \multicolumn{11}{l}{1$^{\rm st}$: Manually-designed ensemble of U-Nets with various backbones.}\\
    \multicolumn{11}{l}{2$^{\rm nd}$: U-Net++ models with manually-reviewed training data.}\\
    \multicolumn{11}{l}{3$^{\rm rd}$: Manually-designed ensemble of U-Nets with different pre-trained encoders.}\\
    \multicolumn{11}{l}{107$^{\rm th}$: Baseline model (U-Net with pre-trained ResNet-34 used as the encoder).}\\
    \multicolumn{11}{l}{S-2: Built-in S-2 cloud detection (thresholding of the cirrus B10 band).}
\end{tabular}
\end{table}

In Table~\ref{tab:results_challenge}, we gather the results obtained by the participating teams over the test S-2 patches (note that other participants might have used other S-2 bands too). We can observe that the differences between our nnU-Nets (with manual post processing applied) and the top 3 teams amount to 0.015, 0.015, and 0.014, respectively. It is of note that all the top-ranked techniques were manually-crafted to this challenge (and underlying data), either at the architectural level (rank 1 and 3), or also at the data level (rank 2), where the authors did review the training patches in a semi-automated way to remove the patches for which the GT quality was questionable. On the other hand, the difference in \Jaccard\, is indeed more visible between the baseline model (ranked 107) and all of the aforementioned ones---it was 0.082 and 0.067 for the winners and our solution.

To verify if small differences (for 38-C, the difference in \Jaccard\, between Cloud-Net and the nnU-Nets was 0.029, whereas for the OCN test data it was 0.015 between the winners and our approach) are possible to spot by the naked eye, we conducted the mean opinion score (MOS) experiment. The responders were given 15 images from the training OCN data with two cloud masks obtained using the nnU-Nets with and without post processing, and the task was to select the mask which ``\emph{more precisely presents the clouds}''. The participants could say that ``\emph{both masks look equally good to me}'' and ``\emph{none}'' (if neither mask was good enough based on the visual inspection). The masks differed in quality measured by \Jaccard\, (min., avg., median, and max. difference in \Jaccard\, was 0.020, 0.030, 0.027, and 0.047). The nnU-Nets with post processing gave the better \Jaccard\,in 6/15 cases (with the avg. improvement in \Jaccard\, of 0.031), whereas in the remaining 9/15 cases the nnU-Nets without post processing resulted in \Jaccard\, better by 0.029 on average. The background of the participants (109 in total) was diverse---we announced MOS at KP Labs, ESA, but also at the university across the students with no remote sensing experience. 

\begin{table}[ht!]
\centering
\scriptsize
  \caption{Average percentage of MOS responses indicating that mask A/B (without/with post processing) was selected as better, and avg. percentage of MOS responses indicating that both masks were equally good or none was good enough according to the responders.}
  \label{tab:mos}
  \renewcommand{\tabcolsep}{2mm}
  \begin{tabular}{cccccc}
    \Xhline{2\arrayrulewidth}
    Images & Mask A (avg. \Jaccard) & Mask B (avg. \Jaccard) & Both & None\\
    \hline
    Higher \Jaccard\, for Mask A & 34.86 (0.485)	& 12.11	(0.456) & 25.32	& 27.73\\
    Higher \Jaccard\, for Mask B & 28.05 (0.698) & 26.58 (0.729)	& 12.87	& 32.52\\
    \hline
    All images & 32.13 (0.570) & 17.90 (0.565) & 20.34	& 29.65\\
    \Xhline{2\arrayrulewidth}
\end{tabular}
\end{table}

\begin{figure}[ht!]
\centering
\scriptsize
\renewcommand{\tabcolsep}{0.2mm}
\newcommand{\mymagnification}{6}
\newcommand{\mywidth}{0.29}
\begin{tabular}{cccc}
\Xhline{2\arrayrulewidth}
     & a) None ($\Delta$: 0.047) & b) Without PP ($\Delta$: 0.021) & c) With PP ($\Delta$: 0.022)\\
     \hline
     \rotatebox{90}{~RGB} & \begin{tikzpicture}
        [,spy using outlines={circle,pink,magnification=\mymagnification,size=1.5cm, connect spies}]
        \node {\pgfimage[width=\mywidth\columnwidth]{"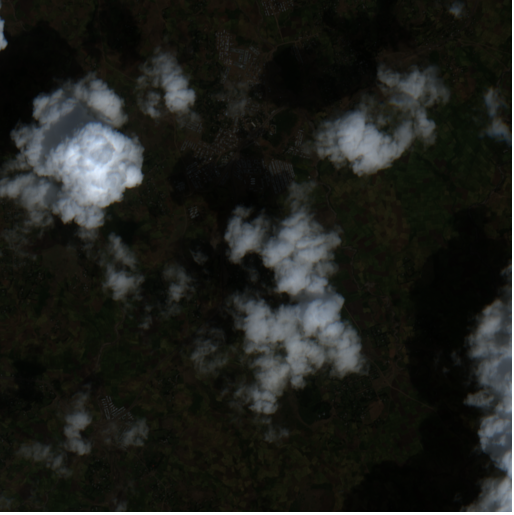"}};
        \end{tikzpicture}& 
        \begin{tikzpicture}
        [,spy using outlines={circle,pink,magnification=\mymagnification,size=1.5cm, connect spies}]
        \node {\pgfimage[width=\mywidth\columnwidth]{"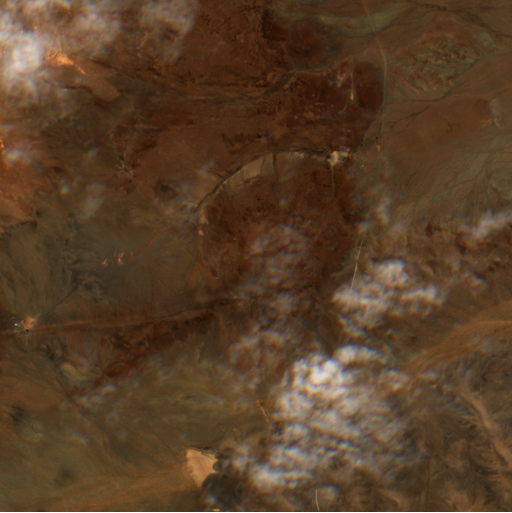"}};
        \end{tikzpicture}& 
        \begin{tikzpicture}
        [,spy using outlines={circle,pink,magnification=\mymagnification,size=1.5cm, connect spies}]
        \node {\pgfimage[width=\mywidth\columnwidth]{"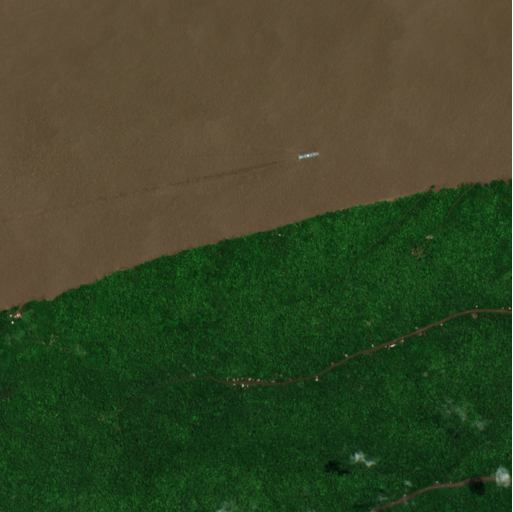"}};
        \spy[every spy on node/.append style={thick},every spy in node/.append style={thick}] on (0.25,0.5) in node [left] at (0.22,-0.5);
        \end{tikzpicture}\\[-0.1cm]
     \hline
     \rotatebox{90}{~Ground truth} &\begin{tikzpicture}
        [,spy using outlines={circle,pink,magnification=\mymagnification,size=1.5cm, connect spies}]
        \node {\pgfimage[width=\mywidth\columnwidth]{"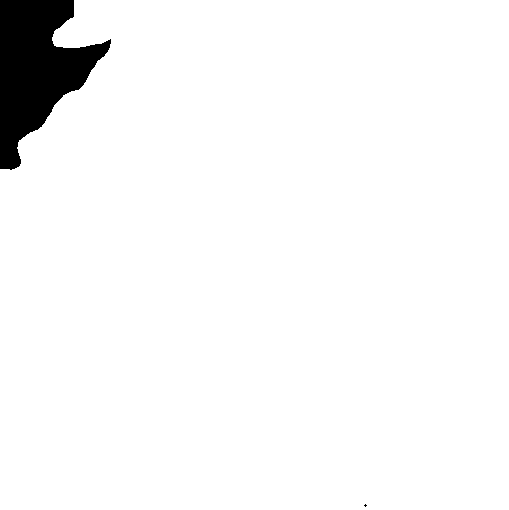"}};
        \end{tikzpicture}& \begin{tikzpicture}
        [,spy using outlines={circle,pink,magnification=\mymagnification,size=1.5cm, connect spies}]
        \node {\pgfimage[width=\mywidth\columnwidth]{"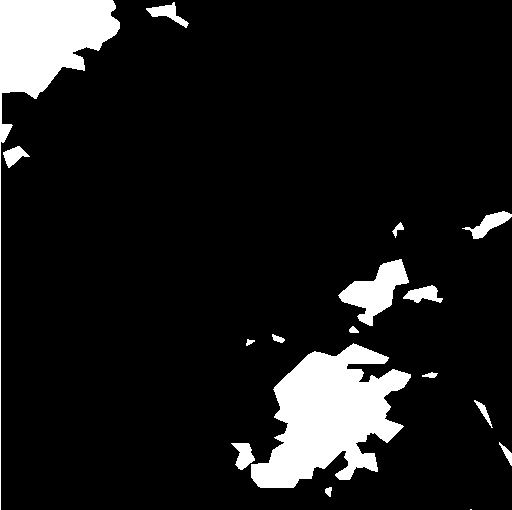"}};
        \end{tikzpicture}& \begin{tikzpicture}
        [,spy using outlines={circle,pink,magnification=\mymagnification,size=1.5cm, connect spies}]
        \node {\pgfimage[width=\mywidth\columnwidth]{"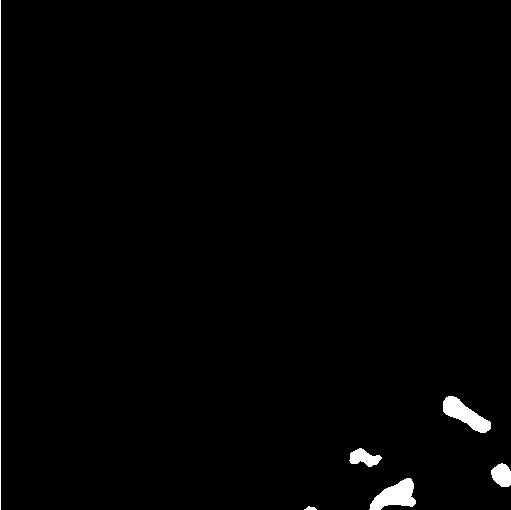"}};
        \spy[every spy on node/.append style={thick},every spy in node/.append style={thick}] on (0.25,0.5) in node [left] at (0.22,-0.5);
        \end{tikzpicture}\\[-0.1cm]
     \hline
     \rotatebox{90}{~Without PP} &\begin{tikzpicture}
        [,spy using outlines={circle,pink,magnification=\mymagnification,size=1.5cm, connect spies}]
        \node {\pgfimage[width=\mywidth\columnwidth]{"S2_examples_12_no"}};
        \end{tikzpicture}&
     \begin{tikzpicture}
        [,spy using outlines={circle,pink,magnification=\mymagnification,size=1.5cm, connect spies}]
        \node {\pgfimage[width=\mywidth\columnwidth]{"S2_examples_6_no"}};
        \end{tikzpicture}&
     \begin{tikzpicture}
        [,spy using outlines={circle,pink,magnification=\mymagnification,size=1.5cm, connect spies}]
        \node {\pgfimage[width=\mywidth\columnwidth]{"S2_examples_9_no"}};
        \spy[every spy on node/.append style={thick},every spy in node/.append style={thick}] on (0.25,0.5) in node [left] at (0.22,-0.5);
        \end{tikzpicture}\\[-0.1cm]
        \hline
     \rotatebox{90}{~With PP} &\begin{tikzpicture}
        [,spy using outlines={circle,pink,magnification=\mymagnification,size=1.5cm, connect spies}]
        \node {\pgfimage[width=\mywidth\columnwidth]{"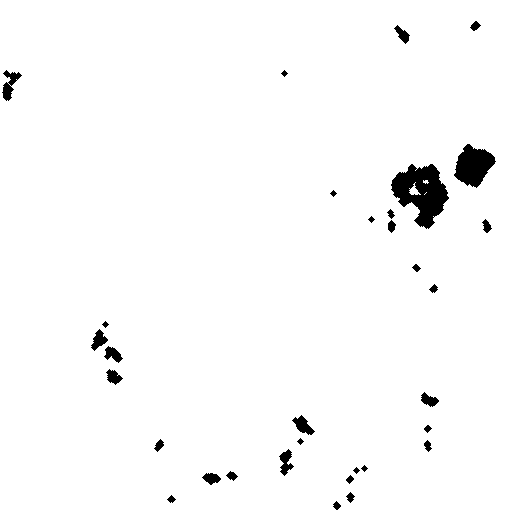"}};
        \end{tikzpicture}&
     \begin{tikzpicture}
        [,spy using outlines={circle,pink,magnification=\mymagnification,size=1.5cm, connect spies}]
        \node {\pgfimage[width=\mywidth\columnwidth]{"S2_examples_6_yes"}};
        \end{tikzpicture}&
     \begin{tikzpicture}
        [,spy using outlines={circle,pink,magnification=\mymagnification,size=1.5cm, connect spies}]
        \node {\pgfimage[width=\mywidth\columnwidth]{"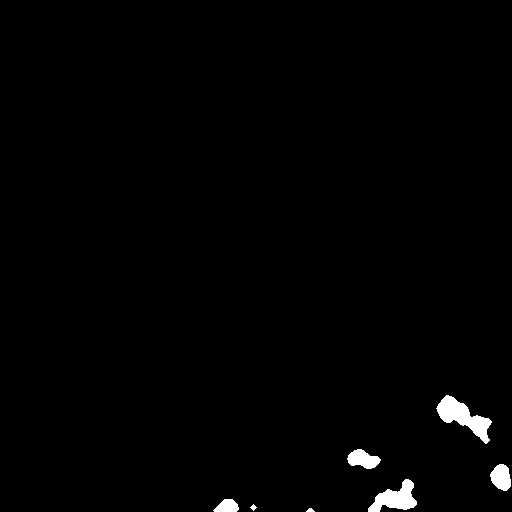"}};
        \spy[every spy on node/.append style={thick},every spy in node/.append style={thick}] on (0.25,0.5) in node [left] at (0.22,-0.5);
        \end{tikzpicture}\\[-0.1cm]
     \Xhline{2\arrayrulewidth}
\end{tabular}
\caption{Examples of cloud masks for which the majority of responders (79.2\%, 63.4\%, and 89.1\%) said that a)\,none of the masks is of enough quality (\Jaccard: 0.894, \Jaccard$^{\rm PP}$: 0.941), b)\,mask obtained without post processing (PP) is better (\Jaccard: 0.594, \Jaccard$^{\rm PP}$: 0.614), and c)~the mask with PP is better (\Jaccard: 0.671, \Jaccard$^{\rm PP}$: 0.693). We also show $\Delta={\rm \Jaccard}^{\rm PP}-{\rm \Jaccard}$.}\label{fig:mos}
\end{figure}
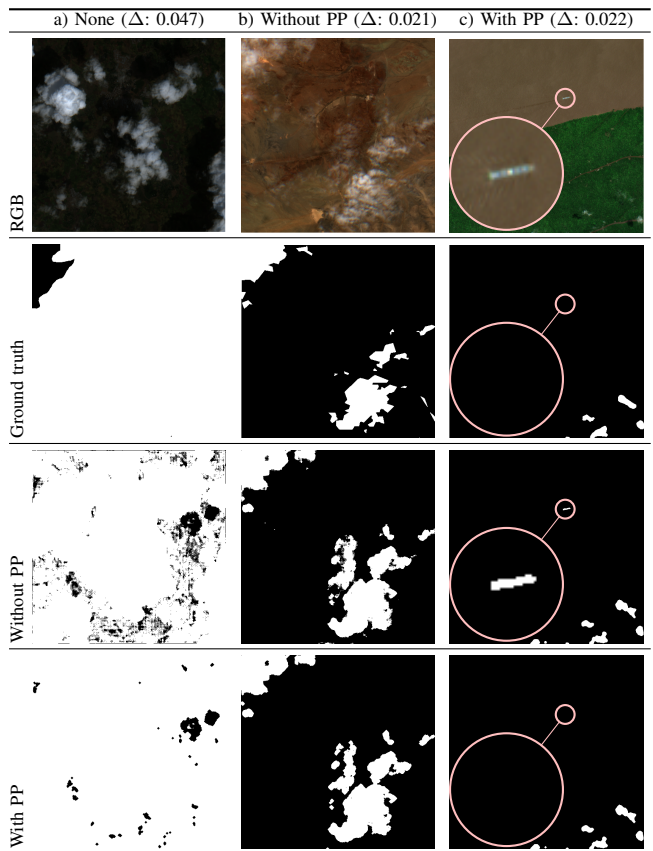

The MOS results (Table~\ref{tab:mos}) indicate a strong disagreement between the responders in all scenarios, i.e.,~across the images for which the mask A or B (without and with post processing) had a larger \Jaccard, and across all (15) images. Although indeed the majority of participants annotated mask A as better than B with a significant margin in the first case (larger \Jaccard\, for mask A, first row), it is not that evident in the second scenario (larger \Jaccard\, for mask B, second row). In all scenarios, approximately half of the responders did not select a single mask as better (the sum of ``Both'' and ``None'' were 53.06\%, 45.38\%, and 49.99\%) which shows that spotting a difference across the cloud masks was challenging to humans, or both masks were perceived as the one of insufficient quality. Further qualitative analysis in Fig.~\ref{fig:mos} highlights the questionable quality of the GT (Fig.~\ref{fig:mos}a) which may notably affect the training process---although \Jaccard\, is large in this case, the cloud mask is unacceptable. In Fig.~\ref{fig:mos}b, the larger \Jaccard\, was in contradiction to the visual investigation (63.4\% participants annotated mask A as better, whereas only 8.9\% selected mask B), and only Fig.~\ref{fig:mos}c presents the scene in which \Jaccard\, was in line with MOS, as a tiny false-positive object was pruned in post processing. We can observe that \Jaccard\, can be misleading, as it is affected by the GT quality. MOS provided the evidence that small differences in \Jaccard\,are often not perceivable (``Both'' in Table~\ref{tab:mos}). Such minor improvements may not be worthwhile in real-world missions due to the development cost/segmentation quality trade-offs. In this case, we believe that the hands-free nnU-Nets are extremely cost-efficient while delivering high-quality performance.

\section{Conclusions and Future Work}

In this letter, we tackled the cloud detection task in a hands-free and fully data-driven manner, and proposed to utilize the nnU-Nets in this context. Although the authors of the nnU-Net framework claimed that it can adapt itself for any biomedical image analysis task, we showed that it can be effectively deployed for Earth observation, and offer the state-of-the-art performance in cloud detection with zero user intervention---to the best of our knowledge, nnU-Nets have never been used in satellite image analysis before. Our experiments, performed over large-scale Landsat-8 and Sentinel-2 sets showed that the suggested segmentation pipeline can be easily used for new image data, and it delivers the cloud detection performance which is on par with or outperforming the hand-crafted algorithms which were manually designed for specific satellite data. Our nnU-Net processing chain was ranked within the top 7\% approaches in the On Cloud N: Cloud Cover Detection Challenge which attracted almost 850 participating teams. 

Our results constitute an interesting departure point for further research. Although the nnU-Nets allowed us to obtain high-quality cloud detection, these models are of large capacity (with up to 30M of trainable parameters for the investigated S-2 set) which influences their size. Hence, it would be difficult to uplink such models trained on the ground to the operational satellite once it is in orbit---this update procedure is being investigated by our team in the Intuition-1 and $\Phi$-sat-2 missions. To make nnU-Nets feasible to be exploited in such a scenario, we are working on the knowledge distillation pipeline that will transform the large-capacity learners (e.g.,~nnU-Nets) into significantly smaller and more compact U-Nets that will later undergo thorough benchmarking~\cite{rs13193981}. Also, we are investigating the robustness of the cloud detection nnU-Nets against different acquisition and noise scenarios through simulating real acquisition conditions in the data-level digital twin approach~\cite{rs13081532}. Finally, selecting an appropriate subset of all training samples can be beneficial to improve the segmentation quality in the most challenging snowy areas~\cite{9554170}.

\ifCLASSOPTIONcaptionsoff
  \newpage
\fi

\bibliographystyle{IEEEtran}
\bibliography{ref_all}

\begin{thebibliography}{10}
\providecommand{\url}[1]{#1}
\csname url@samestyle\endcsname
\providecommand{\newblock}{\relax}
\providecommand{\bibinfo}[2]{#2}
\providecommand{\BIBentrySTDinterwordspacing}{\spaceskip=0pt\relax}
\providecommand{\BIBentryALTinterwordstretchfactor}{4}
\providecommand{\BIBentryALTinterwordspacing}{\spaceskip=\fontdimen2\font plus
\BIBentryALTinterwordstretchfactor\fontdimen3\font minus
  \fontdimen4\font\relax}
\providecommand{\BIBforeignlanguage}[2]{{%
\expandafter\ifx\csname l@#1\endcsname\relax
\typeout{** WARNING: IEEEtran.bst: No hyphenation pattern has been}%
\typeout{** loaded for the language `#1'. Using the pattern for}%
\typeout{** the default language instead.}%
\else
\language=\csname l@#1\endcsname
\fi
#2}}
\providecommand{\BIBdecl}{\relax}
\BIBdecl

\bibitem{JEPPESEN2019247}
J.~Jeppesen \emph{et~al.}, ``A cloud detection algorithm for satellite imagery
  based on deep learning,'' \emph{Remote Sens. Environ.}, vol. 229, pp. 247 --
  259, 2019.

\bibitem{Li2021}
L.~Li \emph{et~al.}, ``A review on deep learning techniques for cloud detection
  methodologies and challenges,'' \emph{Signal, Image and Video Processing},
  vol.~15, no.~7, pp. 1527--1535, 2021.

\bibitem{38-cloud-1}
S.~{Mohajerani} and P.~{Saeedi}, ``{Cloud-Net:} an end-to-end cloud detection
  algorithm for {Landsat 8} imagery,'' in \emph{Proc. IGARSS}, 2019, pp.
  1029--1032.

\bibitem{Mahajan2020}
S.~Mahajan and B.~Fataniya, ``Cloud detection methodologies: variants and
  development---a review,'' \emph{Complex {\&} Intelligent Systems}, vol.~6,
  no.~2, pp. 251--261, 2020.

\bibitem{rs13081532}
J.~Nalepa \emph{et~al.}, ``Towards on-board hyperspectral satellite image
  segmentation: Understanding robustness of deep learning through simulating
  acquisition conditions,'' \emph{Remote Sensing}, vol.~13, no.~8, 2021.

\bibitem{rs13204100}
M.~Domnich \emph{et~al.}, ``{KappaMask: AI-Based Cloudmask Processor for
  Sentinel-2},'' \emph{Remote Sensing}, vol.~13, no.~20, 2021.

\bibitem{38-cloud-2}
S.~Mohajerani \emph{et~al.}, ``A cloud detection algorithm for remote sensing
  images using fully convolutional neural networks,'' in \emph{Proc. MMSP},
  2018, pp. 1--5.

\bibitem{10.1117/1.JRS.15.028507}
M.~Shao and Y.~Zou, ``{Multi-spectral cloud detection based on a
  multi-dimensional and multi-grained dense cascade forest},'' \emph{Journal of
  Applied Remote Sensing}, vol.~15, no.~2, pp. 1 -- 14, 2021.

\bibitem{8898628}
D.~{Varshney} \emph{et~al.}, ``Deep convolutional networks for cloud detection
  using {Resourcesat-2} data,'' in \emph{Proc. IGARSS}, 2019, pp. 9851--9854.

\bibitem{ZHU2015269}
Z.~Zhu \emph{et~al.}, ``{Improvement and expansion of the Fmask algorithm:
  cloud, cloud shadow, and snow detection for Landsats 4–7, 8, and Sentinel 2
  images},'' \emph{Remote Sens. Environ.}, vol. 159, pp. 269--277, 2015.

\bibitem{ZHU201283}
Z.~Zhu and C.~Woodcock, ``Object-based cloud and cloud shadow detection in
  landsat imagery,'' \emph{Remote Sens. Environ.}, vol. 118, pp. 83--94, 2012.

\bibitem{Yanan_2020}
G.~Yanan \emph{et~al.}, ``Cloud detection for satellite imagery using deep
  learning,'' \emph{J. Phys. Conf. Ser.}, vol. 1617, no.~1, p. 012089, 2020.

\bibitem{LI2019197}
Z.~Li \emph{et~al.}, ``Deep learning based cloud detection for medium and high
  resolution remote sensing images of different sensors,'' \emph{ISPRS Journal
  of Photogrammetry and Remote Sensing}, vol. 150, pp. 197--212, 2019.

\bibitem{9615070}
A.~Francis \emph{et~al.}, ``{SEnSeI: A Deep Learning Module for Creating Sensor
  Independent Cloud Masks},'' \emph{IEEE TGRS}, vol.~60, pp. 1--21, 2022.

\bibitem{DBLP:conf/gecco/LorenzoN18}
P.~R. Lorenzo and J.~Nalepa, ``Memetic evolution of deep neural networks,'' in
  \emph{Proc. GECCO}.\hskip 1em plus 0.5em minus 0.4em\relax {ACM}, 2018, pp.
  505--512.

\bibitem{DBLP:conf/gecco/LorenzoNKRP17}
P.~R. Lorenzo \emph{et~al.}, ``Particle swarm optimization for hyper-parameter
  selection in deep neural networks,'' in \emph{Proc. GECCO}, 2017, pp.
  481--488.

\bibitem{DBLP:journals/corr/abs-2112-09245}
X.~Dong \emph{et~al.}, ``Automated deep learning: Neural architecture search is
  not the end,'' \emph{CoRR}, vol. abs/2112.09245, 2021.

\bibitem{10.1007/978-3-030-86517-7_28}
P.~Salinas \emph{et~al.}, ``{Automated Machine Learning for Satellite Data:
  Integrating Remote Sensing Pre-trained Models into AutoML Systems},'' in
  \emph{Proc. ECML PKDD}.\hskip 1em plus 0.5em minus 0.4em\relax Springer,
  2021, pp. 447--462.

\bibitem{isensee_nnu-net_2021}
F.~Isensee \emph{et~al.}, ``\BIBforeignlanguage{en}{{nnU}-{Net}: a
  self-configuring method for deep learning-based biomedical image
  segmentation},'' \emph{\BIBforeignlanguage{en}{Nature Methods}}, vol.~18,
  no.~2, pp. 203--211, Feb. 2021.

\bibitem{DIAZ202125}
O.~Diaz \emph{et~al.}, ``{Data preparation for artificial intelligence in
  medical imaging: A comprehensive guide to open-access platforms and tools},''
  \emph{Physica Medica}, vol.~83, pp. 25--37, 2021.

\bibitem{rs13193981}
M.~Ziaja \emph{et~al.}, ``Benchmarking deep learning for on-board space
  applications,'' \emph{Remote Sensing}, vol.~13, no.~19, 2021.

\bibitem{9554170}
B.~Grabowski \emph{et~al.}, ``Towards robust cloud detection in satellite
  images using {U-Nets},'' in \emph{Proc. IGARSS}, 2021, pp. 4099--4102.

\end{thebibliography}

\end{document}


\begin{center} \textbf{Self-Configuring nnU-Nets Detect Clouds in Satellite Images \\(Supplementary Material)}\end{center}
\begin{center} Bartosz Grabowski, Maciej Ziaja, Michal Kawulok, Nicolas Long{\'e}p{\'e}, Bertrand Le Saux, Jakub Nalepa\end{center}
\begin{center} \texttt{jnalepa@ieee.org}\end{center}

This supplementary material contains the spectral and spatial characteristics of the datasets used in our experiments (Table~\ref{tab:datasets}), together with the distribution of all metrics obtained using the nnU-Nets across all test 38-C scenes (Figure~\ref{fig:38c_violin}). To access the detailed nnU-Net architectures elaborated for both datasets (Landsat-8 and Sentinel-2) together with their parameterization, and our script for transforming the multispectral images into the NIfTI files required by the nnU-Nets, see \url{https://gitlab.com/jnalepa/nnUNets_for_clouds}.

\begin{table}[ht!]
\scriptsize
  \caption{The characteristics of the 38-Cloud (L-8) and OCN (S-2) datasets.}
  \label{tab:datasets}
  \renewcommand{\tabcolsep}{1.8mm}
  \begin{tabular}{ccccccccccc}
    \Xhline{2\arrayrulewidth}
    \multicolumn{2}{c}{Band ID} && \multicolumn{2}{c}{Band name} && \multicolumn{2}{c}{Wavelength [nm]} && \multicolumn{2}{c}{Resolution [m]}\\
    \cline{1-2} \cline{4-5} \cline{7-8} \cline{10-11}
    38-C & OCN && 38-C & OCN && 38-C & OCN && 38-C & OCN \\
    \hline
    B02 & B02 && \multicolumn{2}{c}{Blue} && 450--515 & 458--523 && 30 & 10\\
    B03 & B03 && \multicolumn{2}{c}{Green} && 520--600 & 543--578 && 30 & 10\\
    B04 & B04 &&  \multicolumn{2}{c}{Red} && 630--680 & 650--680 && 30 & 10\\
    B05 & B08 &&  \multicolumn{2}{c}{NIR} && 845--885 & 785--899 && 30 & 10\\
  \Xhline{2\arrayrulewidth}
\end{tabular}
\end{table}

\newcommand{\Jaccard}{JI}
\newcommand{\Precision}{Pr}
\newcommand{\Recall}{Re}
\newcommand{\Specificity}{Spe}
\newcommand{\OverallAccuracy}{OA}

\begin{figure}[ht!]
\centering
\includegraphics[width=0.87\columnwidth]{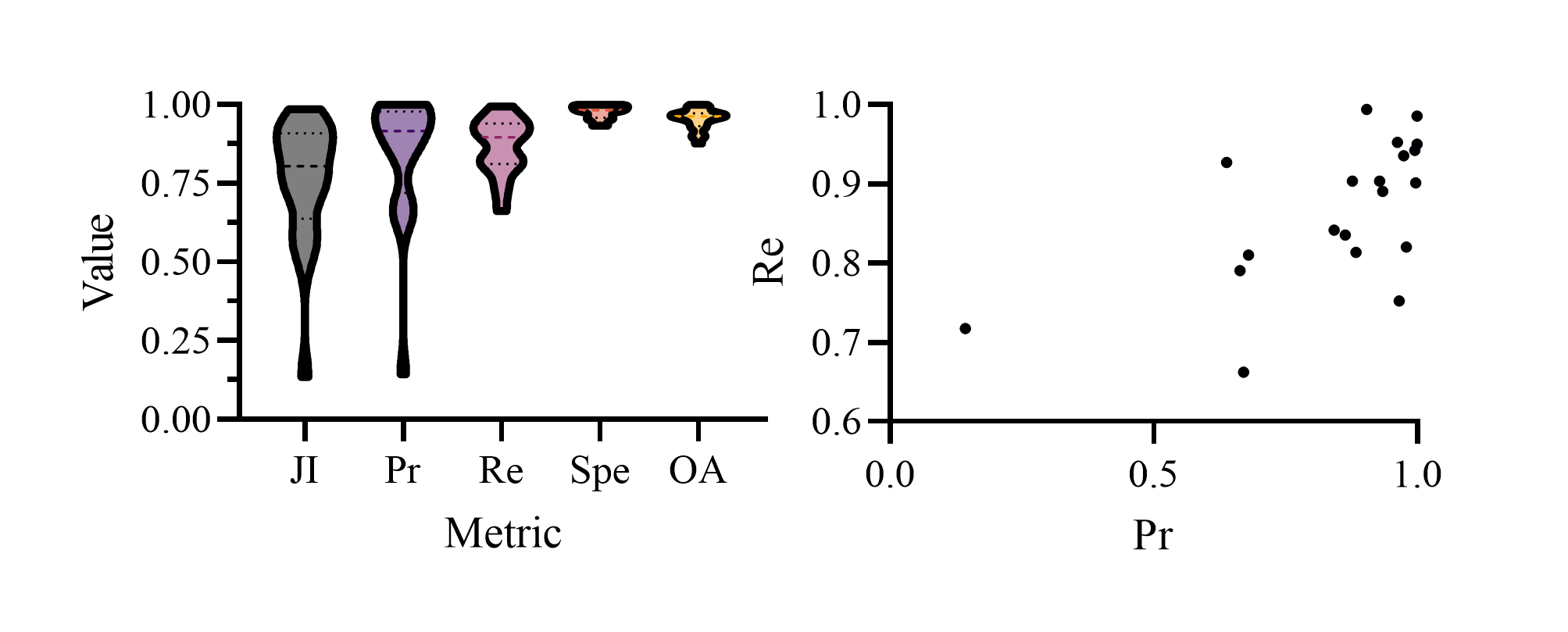}\\[-0.5cm]
\caption{The distribution of all metrics obtained using the nnU-Nets across all test 38-C scenes (left), together with the corresponding precision-recall scatter plot (right). The most challenging scene (\Precision: 0.143 and \Recall: 0.717) presents the mountainous areas for which the nnU-Nets delivered false-positive detections (see the last row in Fig.~2 in the letter). It is of note that the L-8 test set contains the snowy areas which are challenging to segment due to a large number of bright pixels that can be mislabeled as clouds (leading to a significant number of false positives).}\label{fig:38c_violin}
\end{figure}
